\documentclass[10pt,twocolumn]{article}
\usepackage[a4paper, total={7in, 10in}]{geometry}
\usepackage{times}
\usepackage{epsfig}
\usepackage{graphicx}
\usepackage{amsmath}
\usepackage{amssymb}
\usepackage{subfig}

\usepackage[pagebackref=true,breaklinks=true,letterpaper=true,colorlinks,bookmarks=false]{hyperref}

\begin{document}

\title{Style Decomposition for Improved  Neural Style Transfer}

\author{Paraskevas Pegios, Nikolaos Passalis and Anastasios Tefas\\
School of Informatics\\
Aristotle University of Thessaloniki\\
{\tt\small \{ppegiosk, passalis, tefas\}@csd.auth.gr}
}

\maketitle

\begin{abstract}
Universal Neural Style Transfer (NST) methods are capable of performing style transfer of arbitrary styles in a style-agnostic manner via feature transforms in (almost) real-time. Even though their unimodal parametric style modeling approach  has been proven adequate to transfer a single style from relatively simple images, they are usually not capable of effectively  handling more complex styles, producing significant artifacts, as well as reducing the quality of the synthesized textures in the stylized image. To overcome these limitations, in this paper we propose a novel universal NST approach that separately models each sub-style that exists in a given style image (or a collection of style images). This allows for better modeling the subtle style differences within the same style image and then using the most appropriate sub-style (or mixtures of different sub-styles) to stylize the content image.
The ability of the proposed approach to a) perform a wide range of different stylizations using the sub-styles that exist in one style image, while giving the  ability to the user to appropriate \textit{mix} the different sub-styles, b) automatically match the most appropriate sub-style to different semantic regions of the content image, improving existing state-of-the-art universal NST approaches, and c) detecting and transferring the sub-styles from collections of images are demonstrated through extensive experiments.

\end{abstract}

\section{Introduction}

The seminal work of Gatys et al.~\cite{gatys2015neural} opened up a new field called Neural Style Transfer (NST). NST is the process of using Convolutional Neural Networks (CNN) to render an image in different  styles.  NST methods are capable of stylizing a content image by generalizing the abstractions of style, as they are expressed in a given style image, into the original content of the input image. NST is capable of producing spectacular stylized images according to vastly different styles and, as a result, it has become a trending topic that has attracted wide attention from both academia and industry.

Early NST methods achieved quite impressive results, but, at the same time, required a slow iterative optimization step~\cite{gatys2015texture, li2016combining,li2017laplacian,li2017demystifying,patel2015visual, risser2017stable}. Another category of NST approaches, the so-called feed-forward methods, were capable of accelerating NST, without requiring solving a new optimization problem for each input image. However, these methods  are limited to transferring a specific style (or a small number of styles) for which they were pre-trained~\cite{dumoulin2017learned,johnson2016perceptual,li2016precomputed,ulyanov2016texture,ulyanov2017improved}. 
The aforementioned limitations were recently resolved by \textit{universal style transfer} methods that employ  one single trainable model to transfer arbitrary artistic styles via feature manipulations using a shared high-level feature space \cite{huang2017arbitrary,li2017universal,sheng2018avatar,wynen2018unsupervised}.

\begin{figure}
	\subfloat[Content image]{
		\includegraphics[width=0.49\linewidth]{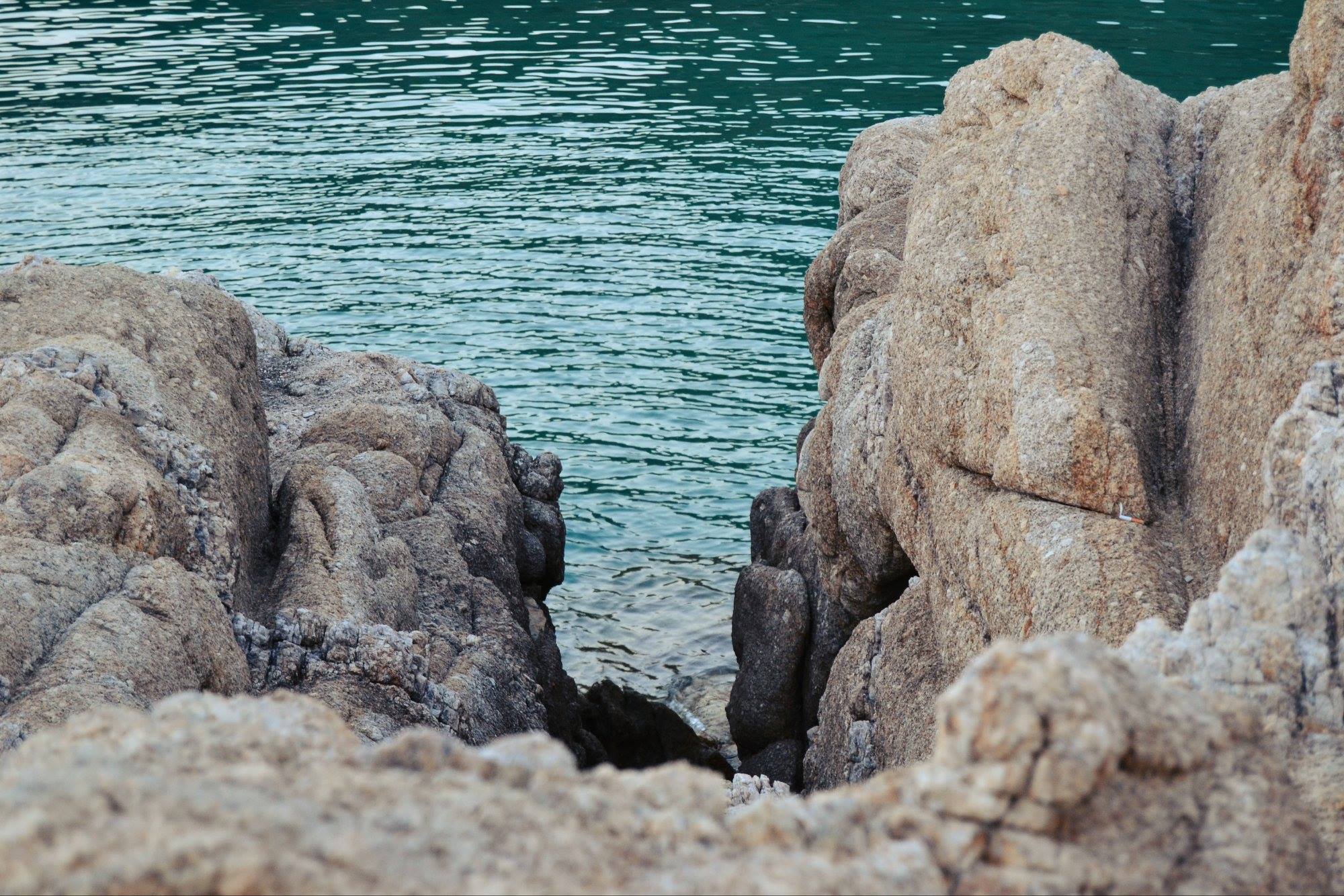}}
	\subfloat[Style image]{	\includegraphics[width=0.49\linewidth]{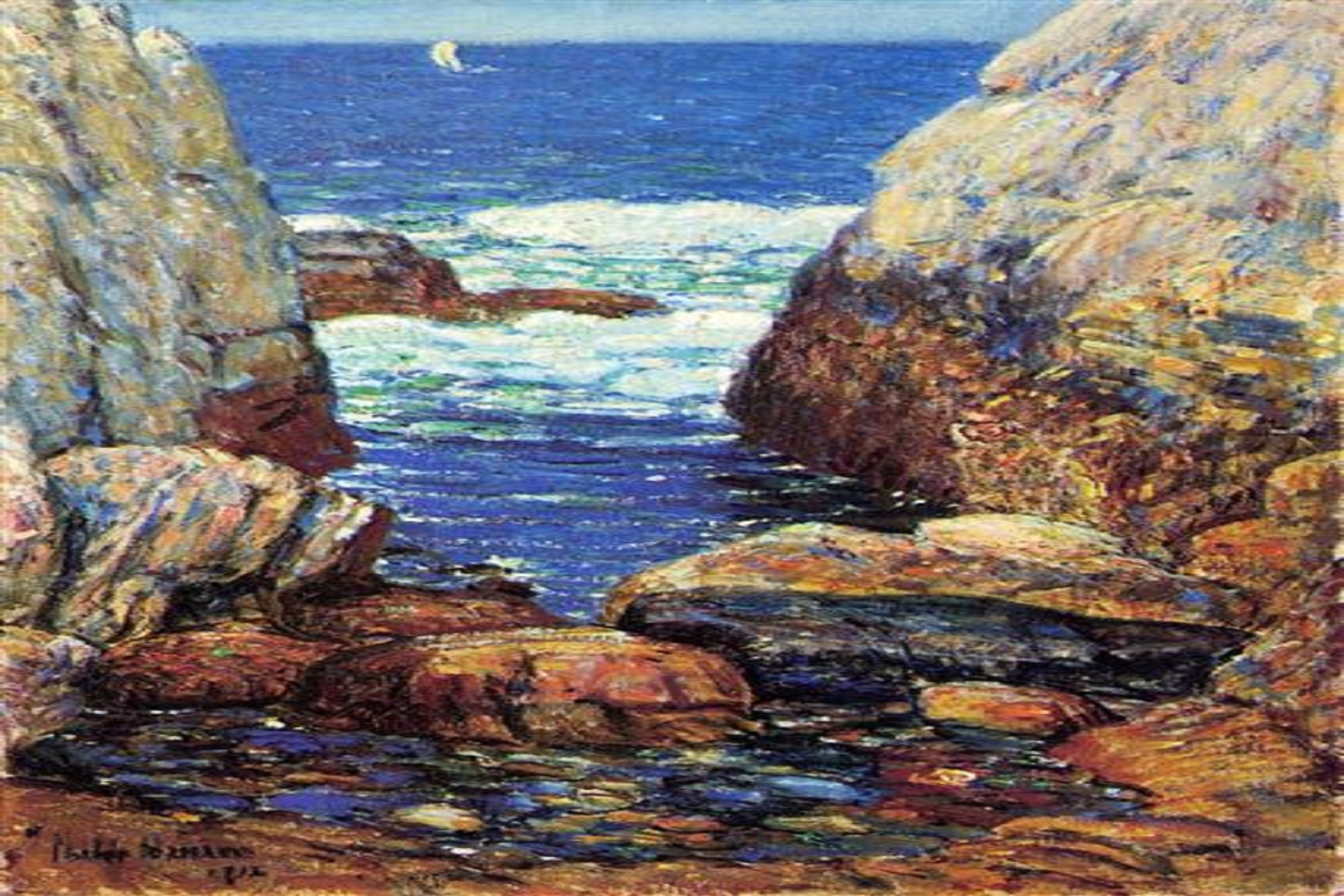}}\\
	
	\subfloat[Improved WCT~\cite{wynen2018unsupervised}]{\label{fig:sample1wct}	
		\includegraphics[width=0.49\linewidth]{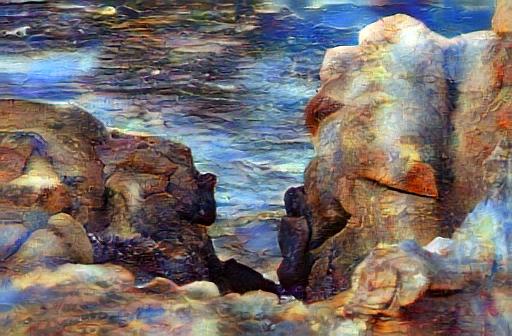}}
	\subfloat[Proposed (Semantic Sub-style Transfer)]{\label{fig:sample1ours}
		\includegraphics[width=0.49\linewidth]{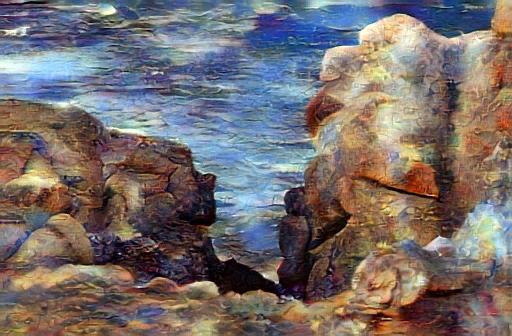}}
	
	\caption{Usually multiple \textit{sub-styles} exist in the same style image. Detecting and separately modeling each sub-style improves the quality of universal style transfer approaches. For example, note the uniform texture of the sea, when the sub-styles are independently modeled and matched to the appropriate regions of the content image (Fig.~\ref{fig:sample1ours}) compared to a state-of-the-art universal style transfer method (Fig.~\ref{fig:sample1wct}) that \textit{leaks} the texture of the rocks into the sea, producing significant artifacts and leading to less detailed texture.
	}
	\label{fig:example1}
\end{figure}

\begin{figure*}
	\centering
	\includegraphics[width=\linewidth]{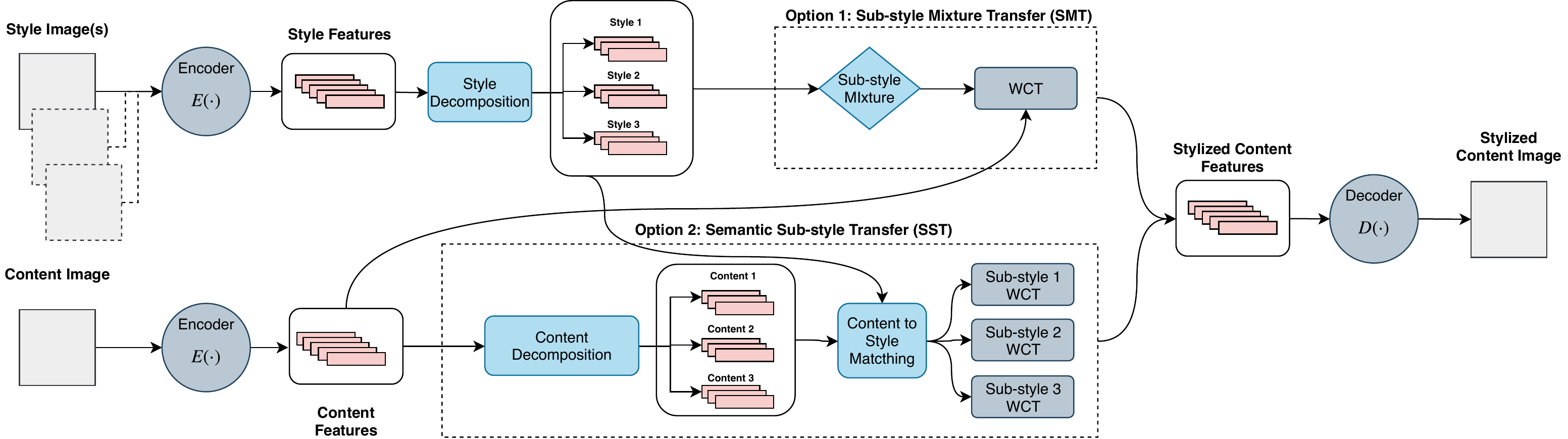}
	\caption{Overview of the proposed style decomposition approach: First, the sub-styles contained in a given style image (or a collection of style images) are detected. Then, two choice exist: a) the user blends the detected sub-styles according the to needs of each application  (SMT) or b) the sub-styles are automatically matched to the most appropriate regions of the content image  (SST). The former provides greater flexibility allowing for producing several different stylized images, while the latter allows for reducing the artifacts in the stylized images and improving the quality of the stylized textures.}
	\label{fig:proposed-method}
\end{figure*}

Even though these universal style transfer approaches  are capable of performing style transfer of arbitrary styles in a style-agnostic manner via feature transforms in (almost) real-time, they usually lead to less impressive results than the other algorithms~\cite{jing2017neural}. Existing universal style transfer approaches assume that the content and style representations extracted by a pre-trained CNN can be described by a single multivariate Gaussian distribution and then perform style transfer by appropriately manipulating the feature statistics of the content image to match those of the target style style. Even though modeling the features using a single Gaussian distribution has been proven adequate to transfer a single style from relatively simple images, this approach is usually not enough to transfer more complex styles from more sophisticated images, e.g., paintings. This is better illustrated in Fig.~\ref{fig:example1}, where a state-of-the-art feature transformation method, that is based on the Whitening Coloring Transformation (WCT)~\cite{li2017universal, wynen2018unsupervised}, was used for transferring the style. Note that the WCT-based method produces artifacts, e.g., the color of the rocks leaks into the sea, due to its inability to adequately model the separate \textit{sub-styles} that exist in the style image. Therefore, we argue that modeling the statistics of the \textit{sub-styles} that exist in an image can improve the quality of neural style transfer. This is illustrated in Fig.~\ref{fig:sample1ours}, where the sub-styles, that were (automatically) detected and modeled, were used to (automatically) stylize the appropriate parts of the content image. Note that the previous artifacts disappeared, while the overall quality of the stylized image improved, e.g., the texture of the rocks is now more detailed.

The main contribution of this paper is the proposal of a feature transformation-based method for universal neural style transfer that separately models each sub-style that exists in a given style image (or a collection of style images), as shown in Fig.~\ref{fig:proposed-method}. This allows for better modeling the subtle style differences within the same style image and then uses the most appropriate sub-style (or mixtures of different sub-styles) to stylize the content image. 

 More specifically, the proposed method is capable of:
\begin{enumerate}
	\item \textbf{Detecting, decomposing and modeling the individual sub-styles contained in a given style image}. Then, the detected sub-styles can be blended according to the needs of the users giving the opportunity to further adjust the process of style transfer. At the same time, it provides a useful tool for discovering and studying the different sub-styles that might exist in a painting. This approach is called \textit{Sub-style Mixture Transfer (SMT)} in this paper. 
	\item \textbf{Automatically matching each detected sub-style to the most appropriate region of content image}. This allows for performing  automatic semantic matching between the detected sub-styles and different regions of the content image, improving the quality of the style transfer and reducing possible artifacts (e.g., as demonstrated in Fig.~\ref{fig:example1}). This approach is called \textit{Semantic Sub-style Transfer (SST)}  in this paper.
	\item \textbf{Detecting and modeling the sub-styles over a collection of input style images}. This method allows for detecting consistent style trends that appear in collections of different images, e.g., detecting consistent painting styles that an artist used through his/her life. This approach is called \textit{Multi-image Sub-style Transfer (MST)}  in this paper.
\end{enumerate}
The ability of the proposed approach to a) perform a wide range of different stylizations using the sub-styles that exist in one style image, while giving the  ability to the user to appropriate \textit{mix} the different sub-styles, b) automatically match the most appropriate sub-style to different semantic regions of the content image, improving the existing state-of-the-art universal NST approaches, and c) detecting and transferring the sub-styles from collections of images are demonstrated though extensive experiments.  Note that the proposed approach is orthogonal to existing NST methods, i.e., it can be readily combined with most of them, providing a powerful framework that can be used towards a better understanding of NST by decomposing the style of images, while giving the opportunity to the users to interfere with the final result, providing a practical NST tool.

The rest of the paper is structured as follows. The related work is discussed in Section~\ref{section:rw}. Then, the proposed style decomposition, mixture and transfer approaches are presented in detail in Section~\ref{section:proposed-method}. Finally, the experimental evaluation of the proposed approach is provided in Section~\ref{section:experimental}, while conclusions are drawn in Section~\ref{section:conclusions}.

\section{Related Work}
\label{section:rw}

The proposed method improves a recently proposed family of NST algorithms, the universal NST methods that employ feature transformations for transferring the style from a given style image into a target content image~\cite{huang2017arbitrary,li2017universal,sheng2018avatar,wynen2018unsupervised}.  The main assumption behind these approaches is that the style is encoded in the feature statistics extracted by a deep CNN. For example, \cite{huang2017arbitrary} transfers the style by transforming the features extracted from the content image using a deep CNN to have the same channel-wise mean and variance as the features of the style image. Then, the stylized image is reconstructed by \textit{decoding} the transformed features. More recently, \cite{li2017universal} employed a more powerful approach aiming to match the covariance of the content features to the covariance of the style features. This significantly improved the quality of NST over the simpler approaches followed by previous works, such as~\cite{huang2017arbitrary}. Note that the stylization process can be controlled by appropriately blending the transformed features with the original ones~\cite{li2017universal, wynen2018unsupervised}, while~\cite{sheng2018avatar} also apply a patch-based feature manipulation module to further improve the quality of the transfer. 

However, these approaches assume that each style image contains only one style and ignores most of the semantic relationships between different sub-styles of the given style image and the corresponding content of the content image, producing artifacts and loosing fine texture details. On contrary, the proposed approach is capable of detecting several sub-styles and matching each sub-style to the most appropriate region of the content image significantly improving the process of style transfer. It is worth noting that the proposed approach is orthogonal to these methods, i.e., it can be readily combined with any of these approaches and further improve the process of style transfer. In this paper, the proposed method is combined with a powerful universal style approach~\cite{li2017universal}, that is based on the Whitening Coloring Transformation (WCT).

There are also some approaches  that focus on building semantic correspondence between the style and the content~\cite{champandard2016semantic, chen2016towards, mechrez2018contextual}. However, these approaches mostly relied on manual semantic annotations for performing the stylization. On the contrary, in this paper, we propose a method that is capable of performing fully automated sub-style detection and decomposition and then matching each semantic region of the content image to the most appropriate sub-style. Furthermore, in~\cite{wynen2018unsupervised}, the style is also transferred from collection of paintings into one target content image. However, the proposed approach is capable of detecting sub-styles even from one single image, without the need of having large collections of paintings of the same artist.

To the best of our knowledge, this is the first unsupervised NST approach that employs style decomposition, in sub-styles that can be extracted from a \textit{single} style image, to improve the process of style transfer by capturing the fine sub-style variations.  The proposed method is able to decompose the content and style automatically, in a fully unsupervised way, and perform semantic-based sub-style transfer, in a style-agnostic manner, in real-time and without any human intervention. Furthermore, whereas previous approaches aimed to merely improve style transfer or to accelerate the process, our approach also leads towards a better understanding of NST by decomposing the style of a single style image. Our approach also gives the opportunity to the users to interfere with the final result or to produce many different outputs with different textures and/or colors from a single style according to their needs.

\section{Proposed Method}
\label{section:proposed-method}

First, the used notation and necessary background are briefly introduced. Then, the proposed methods are derived and described in detail.

\subsection{Background}

Let $E(\cdot)$ denote a CNN encoder that extracts a set of feature maps and  $D(\cdot)$ be a symmetric decoder, which was trained to reconstruct the original image through the features extracted from $E(\cdot)$. Given an input image $\mathbf{I}$ we calculate the summary of first and second order statistics of the vectorized feature map $\mathbf{F} = E(\mathbf{I})$ $\in$ $\mathbb{R}^{C \times HW}$, where $N = HW$ is the number of extracted feature vectors and  $H$ and $W$ are the height and width of the feature map, as follows:
\begin{equation}
\label{eq:0}
\begin{split}
\boldsymbol{\mu}  & = \frac { 1 } { N } \sum _ { i = 1 } ^ { N }  \mathbf { f } _ { i } \in \mathbb{R}^{C}  \\
\boldsymbol{\Sigma} & = \frac { 1 } { N } \sum _ { i = 1 } ^ { N }  \left( \mathbf { f } _ { i } - \boldsymbol { \mu }  \right) \left( \mathbf { f } _ { i } - \boldsymbol { \mu } \right) ^ { \top }  \in \mathbb{R}^{C \times C}
\end{split}
\end{equation}
where $\mathbf{f}_{i} \in \mathbb{R}^C$ is the $i$-th feature vector of the extracted feature maps and $C$ is the number of channels/feature maps.

Given a pair of an input content image $\mathbf{I}_c$ and a style image $\mathbf{I}_s$, we first extract the features $  \mathbf{F}_{c} =  E(\mathbf{I}_c) \in$ $\mathbb{R}^{C \times N}$ and $ \mathbf{F}_{s}  = E(\mathbf{I}_s)  \in \mathbb{R}^{C \times N}$ respectively. In this work, the WCT is employed for transferring the statistics from the style features $\mathbf{F}_{s}$ to the target content image, as proposed in~\cite{li2017universal}:
\begin{equation}
\mathbf{F}_{cs} = \mathbf{E}_{s}\mathbf{D}^{1/2}_{s}\mathbf{E}^T_{s}\mathbf{F}_{cw} + \boldsymbol{\mu}_{s},
\end{equation}
where 
\begin{equation}
\label{eq:6}
\begin{split}
\mathbf{F}_{cw} = \mathbf{E}_{c}\mathbf{D}^{-1/2}_{c}\mathbf{E}^T _{c}(\mathbf{F}_{c}- \boldsymbol{\mu}_{c}),
\end{split}
\end{equation}
and $\mathbf{D}_c$ ($\mathbf{D}_s$) is a diagonal matrix with the eigenvalues of the content (style) covariance matrix $\boldsymbol{\Sigma}_{c} \in  \mathbb{R}^{C\times C}$ ($ \boldsymbol{\Sigma}_{s} \in \mathbb{R}^{C\times C}$) of the centered content (style) feature map $\mathbf{F}_{c} - \boldsymbol{\mu}_{c}$ ($\mathbf{F}_{s} - \boldsymbol{\mu}_{s}$) and $\mathbf{E}_{c}$ ($\mathbf{E}_{s}$) is the corresponding  matrix of eigenvectors. $\mathbf{F}_{cw}$ is the whitened content feature and $\mathbf{F}_{cs}$ is the colored (stylized) output. The additions/subtractions between a matrix and a vector are appropriately broadcasted. The aforementioned whitening-coloring transformation is denoted by $\mathbf{F}_{cs}=WCT (\mathbf{F}_{c}, \mathbf{F}_{s})$. Then, the stylized image $\mathbf{I}_{cs}$ can be readily obtained by inverting the transformed features using the decoder $D(\cdot)$, i.e., $
\mathbf{I}_{cs} = D(\mathbf{F}_{cs})$.
The stylization process can be controlled through a blending parameter $\alpha$ (style weight) allowing for keeping some of the original style:
\begin{equation}
\mathbf{I}_{cs} = D(\alpha \mathbf{F}_{cs} + (1 - \alpha) \mathbf{F}_{c}) 
\end{equation}

To more accurately stylize the input image cascaded autoencoders of various depths can be also used, as suggested in~\cite{li2017universal}. This process is also called \textit{multi-level stylization}. In this case, the transformed features can be blended both with the original content image (using a similarly defined content weight $\delta$), as well as with the (partially) stylized and reconstructed image, as proposed in~\cite{wynen2018unsupervised}.

\subsection{Style Transfer using Sub-style Decomposition}

The method proposed in this paper does not directly transfer the statistics of the whole feature space of the style image into the feature space of the content image. Instead, it first detects and decomposes the sub-styles that exist in the original style image, as shown in Fig.~\ref{fig:proposed-method}. To this end, in this work, we first detect $K_s$ independent style components using Independent Component Analysis~\cite{lee1998independent}. Therefore, the feature space of a style image is represented as a linear combination of $K_s$ independent style sources:
\begin{equation}
\mathbf{F}_s = \mathbf{A} \mathbf{S},
\end{equation}
where $\mathbf{A}\in \mathbb{R}^{C\times K_s}$ is the mixing matrix and the matrix  $\mathbf{S}\in \mathbb{R}^{K_s \times N}$  contains the separated sub-styles signals for each feature vector. Note that given a feature vector $\mathbf{f}_i$ we can readily obtain its sub-styles using the mixing matrix: $\mathbf{s}_i = \mathbf{A}^{-1} \mathbf{f}_i$. Then, we group the decomposed style feature vectors $\mathbf{s}_i$ into $K_s$ clusters that corresponds to the $K_s$ dominating sub-styles. Any clustering algorithm can be used to this end. In this work, we employ a Gaussian Mixture Model to obtain $K_s$ style clusters~\cite{reynolds2015gaussian}. The i-th style cluster is denoted by:
\begin{equation}
\mathcal{S}_i = \{\mathbf{f}_{s,ij}, j= 1 \dots N_{s,i}\},
\end{equation}
where $\mathbf{f}_{s,ij}$ is the $j$-th feature vector extracted from the style image and belongs to the cluster that corresponds to the $i$-th sub-style. The total number of feature vectors of the $i$-th style cluster is denoted by  $N_{s,i}$. Therefore, we can extract the summary statistics for each of detected sub-styles, using the feature vectors that belong to each cluster, as:
\begin{equation}
\label{eq:statistics}
\begin{split}
\boldsymbol{\mu}_{s, i}  & = \frac { 1 } {N_{s,i} } \sum _ { j = 1 } ^ {  N_{s,i} }  \mathbf{f}_{s, ij} \in \mathbb{R}^{C}  \\
\boldsymbol{\Sigma}_{s, i} & = \frac { 1 } {N_{s,i}} \sum _ { j = 1 } ^ {  N_{s,i}}  \left( \mathbf{f}_{s, ij} - \boldsymbol { \mu }_{s, i}  \right) \left( \mathbf{f}_{s, ij} - \boldsymbol { \mu }_{s, i} \right) ^ { \top }  \in \mathbb{R}^{C \times C}
\end{split}.
\end{equation}
That is, the style expressed by the $i$-th style cluster can be described by the mean $\boldsymbol{\mu}_{s, i}$ and covariance $\boldsymbol{\Sigma}_{s, i}$. Note that even though the clustering is performed using the vectors $\mathbf{s}_i$, the summary statistics are calculated in the original feature space, since this is the space used for transferring the statistics into the content features.

After detecting, decomposing and describing each of the detected sub-style several choices exist. The three different approaches that are used in this paper are described below:

\textbf{Sub-style Mixture Transfer (SMT):}
The first one is to allow the user to define a mixture of different sub-styles that he/she want to use for stylizing the content image. The user defines a mixing coefficient $\beta_i \in [0, 1]$ that expresses the influence of the $i$-th detected sub-style in the stylized image. Therefore, after producing the stylized features according to each feature sub-style using WCT, the stylized features are blended as:
\begin{equation}
\mathbf{F}_{cs} = \frac{1}{\sum_{i=1}^{K_s} \beta_i} \sum_{i=1}^{K_s} \beta_i \mathbf{F}_{cs, i},
\end{equation}
where $\mathbf{F}_{cs, i}$ are the transformed feature vectors according to the statistic of the $i$-th style cluster, i.e., using the mean $\boldsymbol{\mu}_{s,i}$ and covariance matrix $\boldsymbol{\Sigma}_{s,i}$. The user might select only one sub-style, or might mix the sub-styles according to the needs of each application. This provides a way to explore the different sub-styles that exist in the same style image and leads towards a better understanding of the artistic styles used in the original style image, as also demonstrated in the experimental evaluation.

\begin{figure}[t]

\centering
	
\subfloat[style]{\includegraphics[width = 0.245\linewidth]{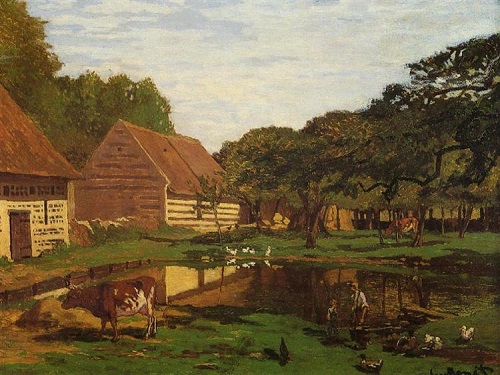}} 
\subfloat[sub-style 1]{\includegraphics[width = 0.245\linewidth]{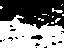}} 
\subfloat[sub-style 2]{\includegraphics[width = 0.245\linewidth]{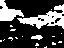}} 
\subfloat[sub-style 3]{\includegraphics[width = 0.245\linewidth]{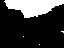}} \\

\subfloat[content]{\includegraphics[width = 0.245\linewidth]{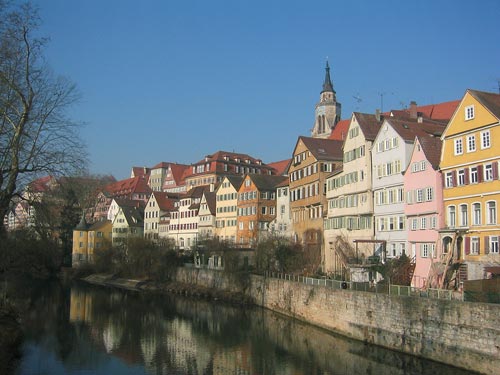}} 
\subfloat[sub-content 1]{\includegraphics[width = 0.245\linewidth]{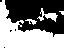}} 
\subfloat[sub-content 2]{\includegraphics[width = 0.245\linewidth]{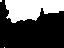}} 
\subfloat[sub-content 3]{\includegraphics[width = 0.245\linewidth]{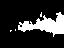}}

\caption{Style and content decomposition. The left column illustrates the original style and content images, while the following three columns depict the three binary masks that correspond to the first, second and third sub-styles/sub-contents detected. Note that regions of the images that have similar style and/or content are grouped together, e.g., the sky, trees, etc.}
\label{fig:masks}
\end{figure}

\textbf{Semantic Sub-style Transfer (SST):}
In the previous approach, the (mixture of) sub-styles were uniformly applied to whole input content image. However, this can lead to artifacts, as it was briefly discussed in the introduction and demonstrated in Fig.~\ref{fig:example1}. To avoid this, we propose matching each sub-style to the most appropriate region of the content image and stylizing the corresponding region using the specific sub-style. To this end, the content image is first segmented into a number of semantic regions using a Gaussian Mixture Model~\cite{reynolds2015gaussian}. This leads into forming a number of clusters $K_c$, each one corresponding to some regions of the input image. Therefore, the $i$-th content cluster is defined as:
\begin{equation}
\mathcal{C}_i = \{\mathbf{f}_{c, ij}, j= 1 \dots N_{c,i}\},
\end{equation}
where $\mathbf{f}_{c, ij}$ is the $j$-th feature vector extracted from the content image and belongs to the cluster that corresponds to the $i$-th sub-content. Again, the total number of feature vectors of the $i$-th cluster is denoted by  $N_{c, i}$. This allows for segmenting both the content and style images into a number of regions, according to their sub-style and sub-content, as shown in Fig.~\ref{fig:masks}. The mean $\boldsymbol{\mu}_{c, i}$ of each sub-content can be similarly calculated as in~(\ref{eq:statistics}). Note that we want to transfer the sub-style between regions with similar style and semantics. For example, in Fig.~\ref{fig:masks} we would like to transfer the style of the sky from the style image into the sky region of the content image and the style of the trees into the trees depicted in the content image. To achieve this, we propose calculating the similarity between each sub-content to each of the sub-styles using the cosine similarity:
\begin{equation}
\label{eq:cosine}
d_{ij} = \frac{{\boldsymbol{\mu}_{c, i}}^T \boldsymbol{\mu}_{s,i}}{||\boldsymbol{\mu}_{c, i}||_2 ||\boldsymbol{\mu}_{s, i}||_2},
\end{equation}
where $||\cdot||_2$ denotes the $l_2$ norm of a vector. Then, we assign each sub-content to the most similar sub-style, as calculated in (\ref{eq:cosine}). This \textit{nearest neighbor matching} approach allows for performing an automatic semantic matching between sub-styles and sub-contents, given that the feature space formed by the CNN encodes part of the semantic information regarding the corresponding regions. Finally, the feature vectors of each sub-content are appropriately transformed using \textit{only} the statistics of the matched sub-style. That is, if the $i$-th sub-style cluster $\mathcal{S}_i$ is matched to the $j$-th sub-content cluster $\mathcal{C}_j$, then the features of $\mathcal{C}_i$ as stylized as $\mathbf{F}_{\mathcal{C}_j s} = WCT(\mathbf{F}_{\mathcal{S}_i}, \mathbf{F}_{\mathcal{C}_j})$, where the notation $\mathbf{F}_{\mathcal{S}_i}$ ( $\mathbf{F}_{\mathcal{C}_j}$) is used to refer to the matrix that contains all the features of the corresponding clusters.  This allows for performing fine-grained stylization, taking into account the semantic content of the content image, significantly improving the stylization process, as demonstrated in the experimental evaluation.

\textbf{Multi-image Sub-style Transfer (MST):} The proposed approach can be also readily applied for transferring the sub-styles from a collection of images, instead of a sole style image. In this case, the feature vectors extracted from multiple images are concatenated and the sub-styles are detected and extracted, as described before. Furthermore, using the proposed semantic matching approach, the sub-styles calculated from a collection of images can be automatically matched to the most appropriate sub-content of the target content image. Apart from improving the quality of style transfer, this approach also provides a useful tool for detecting and transferring consistent sub-styles that arise in a collection of images. For example, if this approach is applied in a collection of painting of the same artist, then the evolution of his/her painting style can be visualized, allowing for better analyzing the paintings of the collection.

\textit{Multi-level Stylization:} Cascaded autoencoding and stylization levels, i.e., multi-level stylization, can be also utilized for the stylization process, as proposed in~\cite{li2017universal}, in order to better capture various levels of details, instead of applying the stylization process only one level. The proposed approach can be also combined with multi-level stylization either by performing the proposed sub-style transfer only on a specific convolutional layer (and using the regular WCT approach on the others~\cite{li2017universal}), or by using the same sub-style and sub-content regions for calculating the feature statistics and transferring the same sub-styles across all the convolutional layers employed for the NST. In this work, the first approach was utilized, since it was proven to be simple and effective.

\graphicspath{{figs/}}

\begin{figure*}[ht]
	\centering

	\subfloat{\includegraphics[width = 0.19\linewidth]{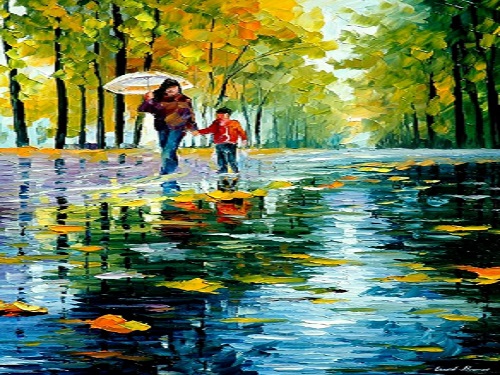}} 	\hspace{0.1px}
	\subfloat{\includegraphics[width = 0.19\linewidth]{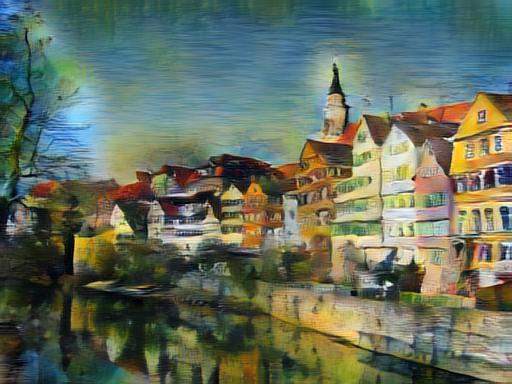}} \hspace{0.1px}
	\subfloat{\includegraphics[width = 0.19\linewidth]{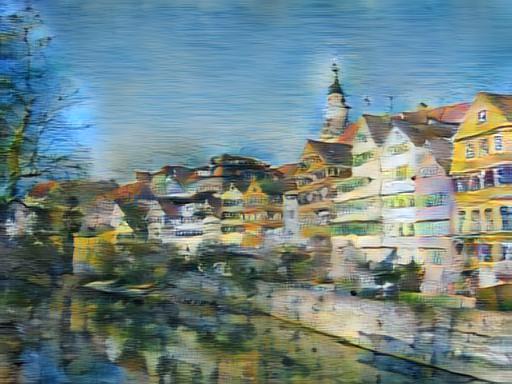}}  \hspace{0.1px}
	\subfloat{\includegraphics[width = 0.19\linewidth]{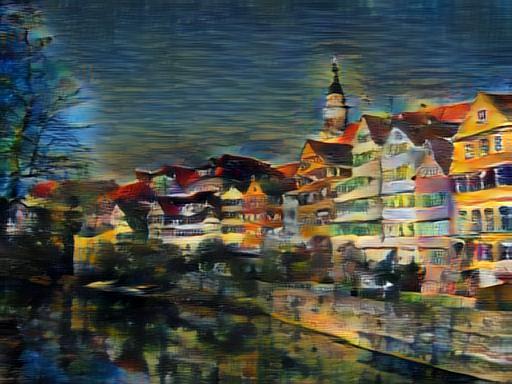}}  \hspace{0.1px}
	\subfloat{\includegraphics[width = 0.19\linewidth]{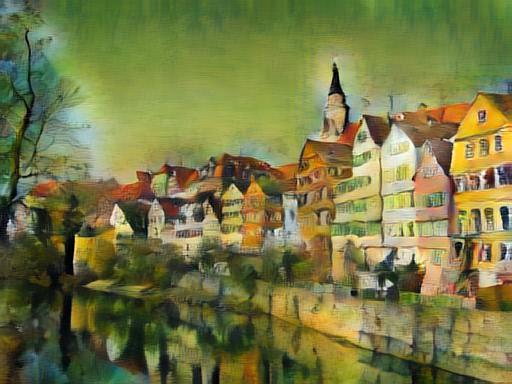}}  \hspace{0.1px} \\
	
	\vspace{-0.8em}
	\subfloat{\includegraphics[width = 0.19\linewidth]{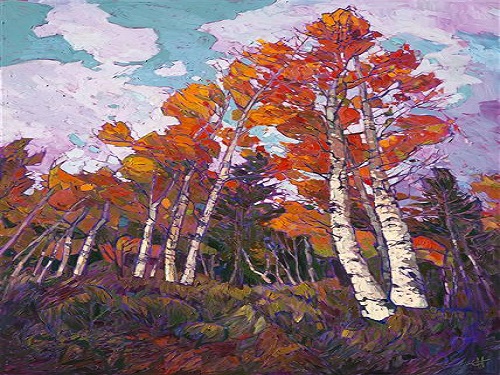}} \hspace{0.1px}
	\subfloat{\includegraphics[width = 0.19\linewidth]{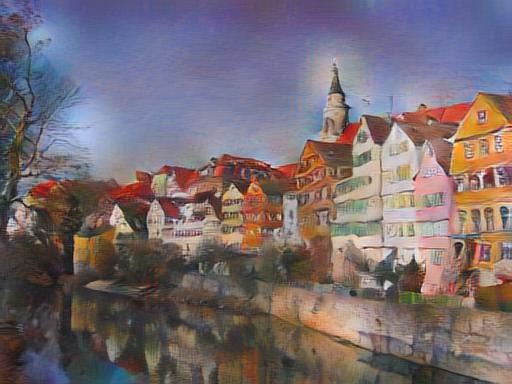}}	 \hspace{0.1px}
	\subfloat{\includegraphics[width = 0.19\linewidth]{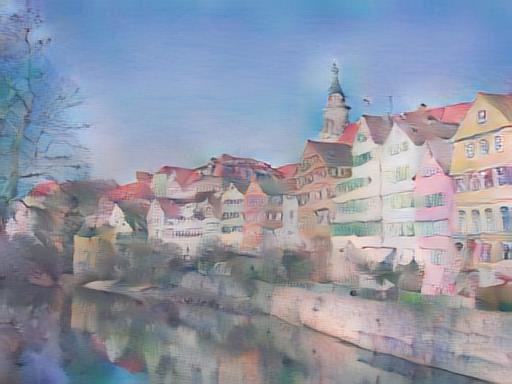}} \hspace{0.1px}
	\subfloat{\includegraphics[width = 0.19\linewidth]{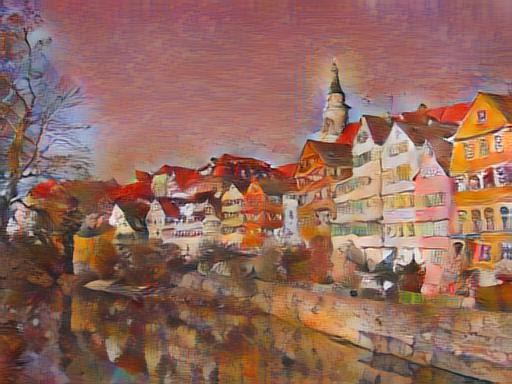}} \hspace{0.1px}
	\subfloat{\includegraphics[width = 0.19\linewidth]{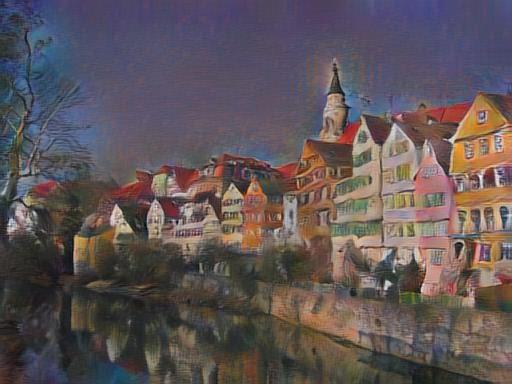}}\hspace{0.1px} \\
	\vspace{-0.8em}
	\subfloat{\includegraphics[width = 0.19\linewidth]{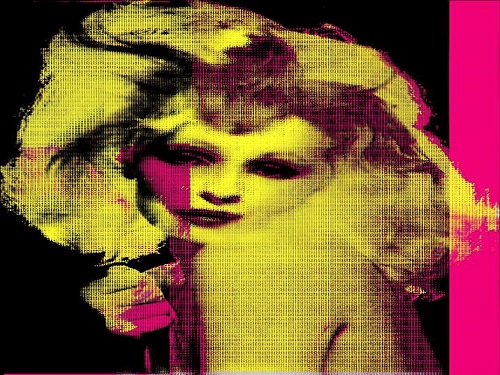}} \hspace{0.1px}
	\subfloat{\includegraphics[width = 0.19\linewidth]{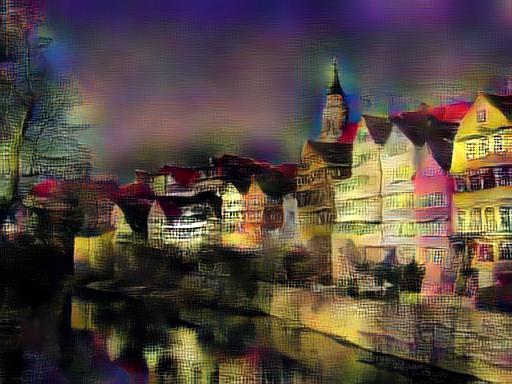}} \hspace{0.1px}	
	\subfloat{\includegraphics[width = 0.19\linewidth]{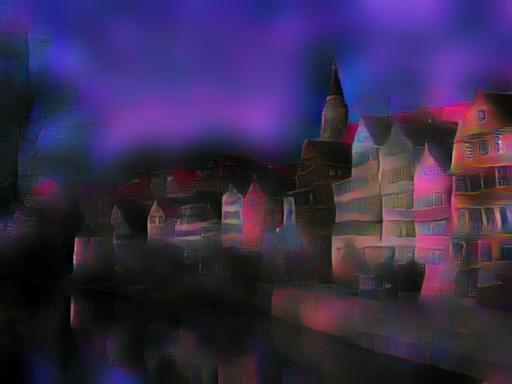}} \hspace{0.1px}
	\subfloat{\includegraphics[width = 0.19\linewidth]{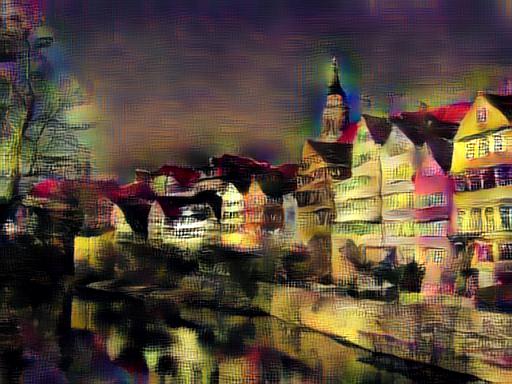}} \hspace{0.1px}
	\subfloat{\includegraphics[width = 0.19\linewidth]{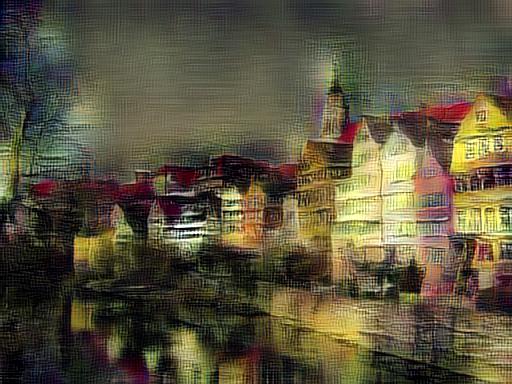}} \hspace{0.1px}
\\
	\vspace{-0.8em}
	\subfloat{\includegraphics[width = 0.19\linewidth]{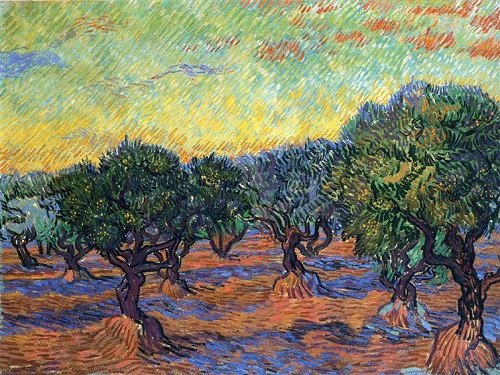}} \hspace{0.1px}
	\subfloat{\includegraphics[width = 0.19\linewidth]{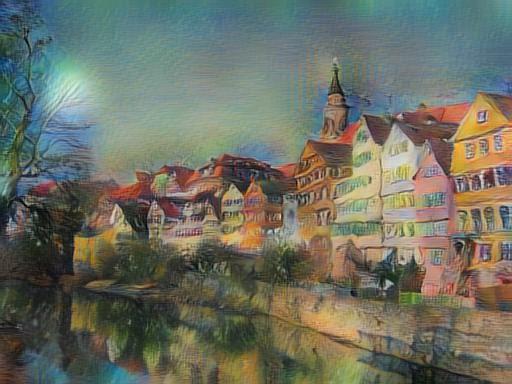}} \hspace{0.1px}	
	\subfloat{\includegraphics[width = 0.19\linewidth]{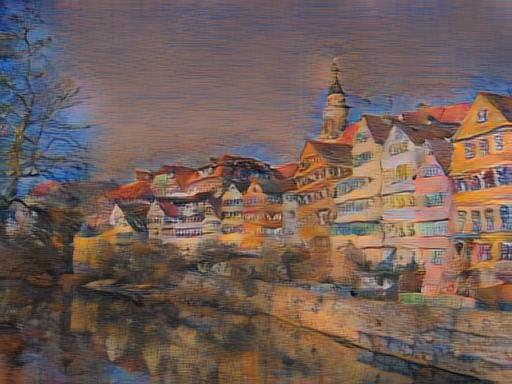}} \hspace{0.1px}
	\subfloat{\includegraphics[width = 0.19\linewidth]{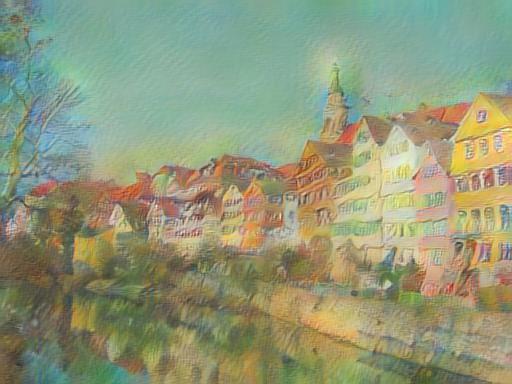}} \hspace{0.1px}
	\subfloat{\includegraphics[width = 0.19\linewidth]{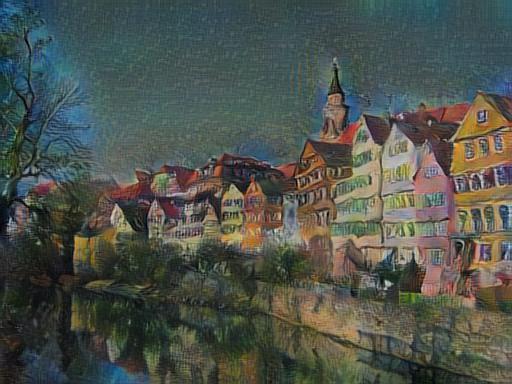}} \hspace{0.1px}
 \\
	\vspace{-0.8em}
	\setcounter{subfigure}{0}
	\subfloat[Style image]{\includegraphics[width = 0.19\linewidth]{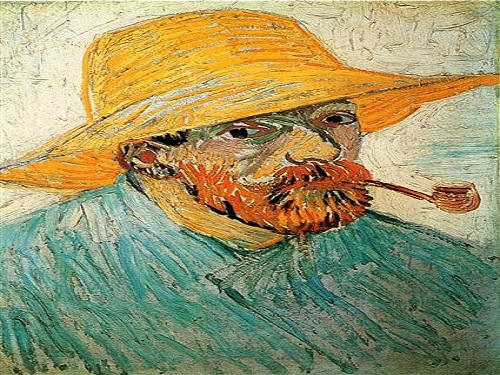}} \hspace{0.1px}
	\subfloat[WCT]{\includegraphics[width = 0.19\linewidth]{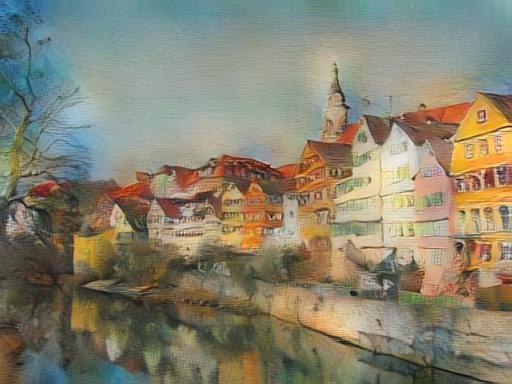}} \hspace{0.1px}	
	\subfloat[SMT (sub-style 1)]{\includegraphics[width = 0.19\linewidth]{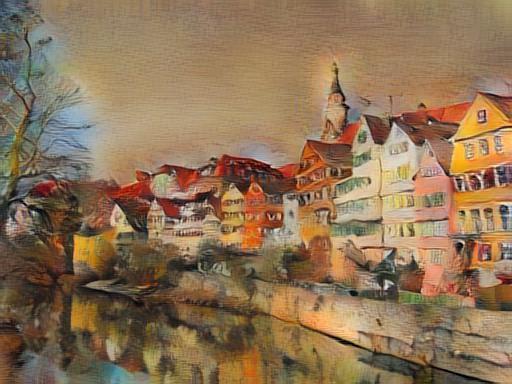}} \hspace{0.1px}
	\subfloat[SMT (sub-style 2)]{\includegraphics[width = 0.19\linewidth]{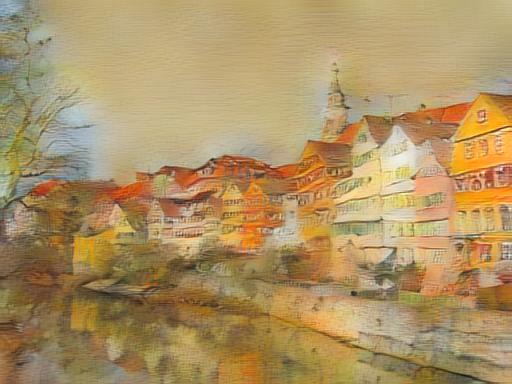}} \hspace{0.1px}
	\subfloat[SMT (sub-style 3)]{\includegraphics[width = 0.19\linewidth]{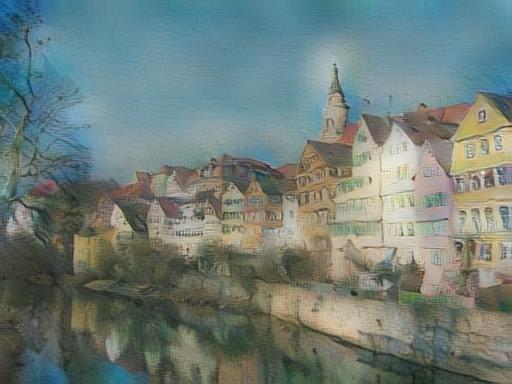}} \hspace{0.1px}
\\
	
	\caption{Sub-style mixture transfer (SMT) examples using features extracted from the fourth pooling layer of the VGG-19 network. To demonstrate the different sub-styles that were detected, the SMT method was used to separately transfer each sub-style.  The blending parameter was set to $\alpha=0.6$ for all the conducted experiments and the same content image was used.  }
	\label{fig:sub-style transfer}
\end{figure*}

\begin{figure*}[ht]
	\centering
	\subfloat{\includegraphics[width = 0.165\linewidth]{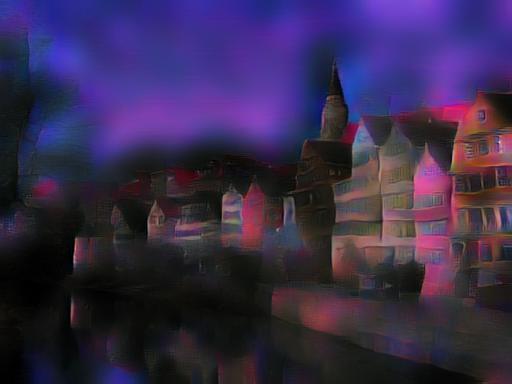}} 
	\subfloat{\includegraphics[width = 0.165\linewidth]{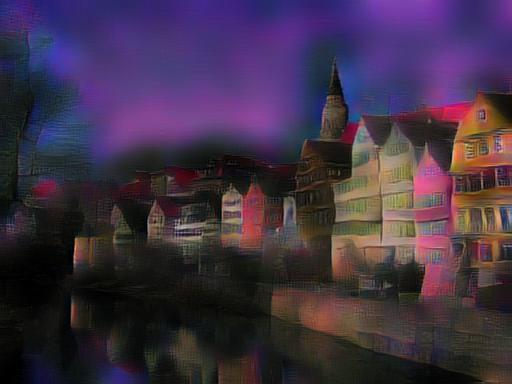}} 
	\subfloat{\includegraphics[width = 0.165\linewidth]{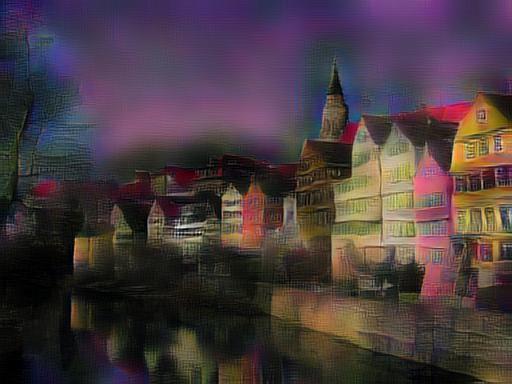}} 
	\subfloat{\includegraphics[width = 0.165\linewidth]{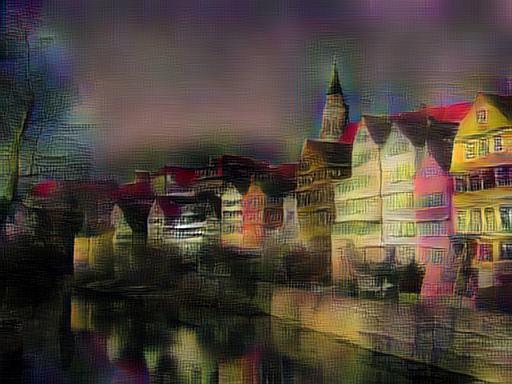}} 
	\subfloat{\includegraphics[width = 0.165\linewidth]{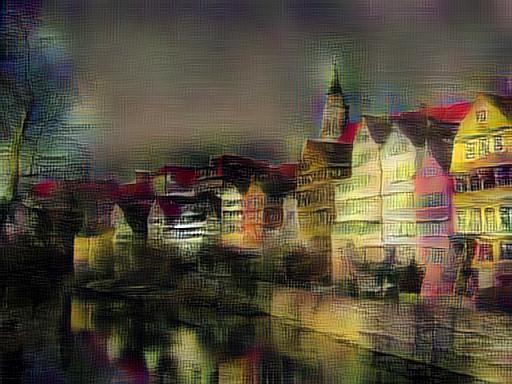}} 
	\subfloat{\includegraphics[width = 0.165\linewidth]{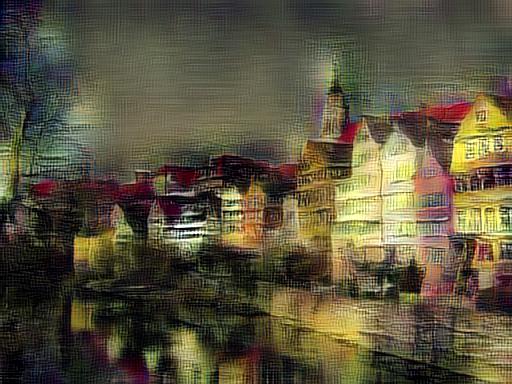}}\\
		\vspace{-0.8em}
	\subfloat{\includegraphics[width = 0.165\linewidth]{interA_I4_substyle_0.jpg}} 
	\subfloat{\includegraphics[width = 0.165\linewidth]{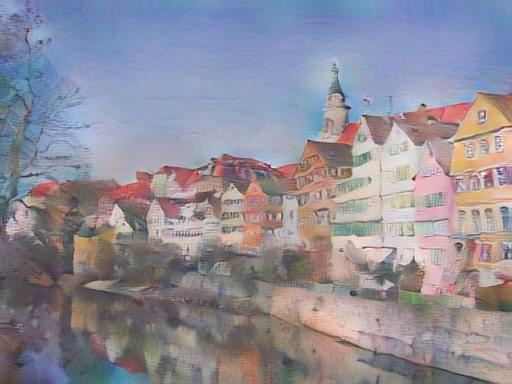}} 
	\subfloat{\includegraphics[width = 0.165\linewidth]{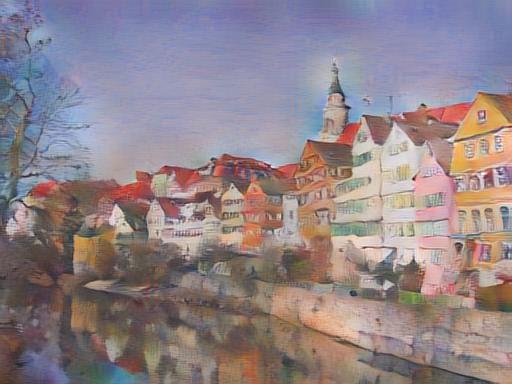}} 
	\subfloat{\includegraphics[width = 0.165\linewidth]{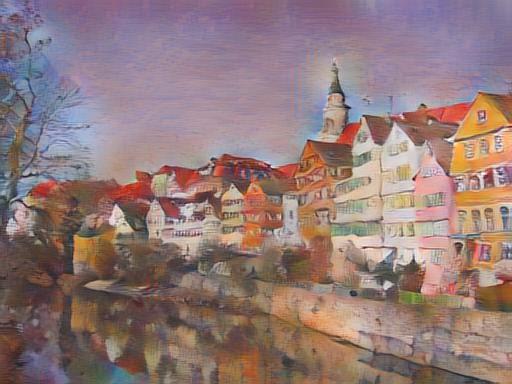}} 
	\subfloat{\includegraphics[width = 0.165\linewidth]{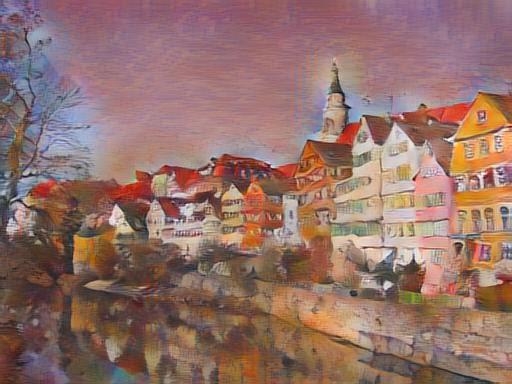}} 
	\subfloat{\includegraphics[width = 0.165\linewidth]{interA_I4_substyle_1.jpg}}\\

	\caption{Sub-style mixture transfer (SMT) by interpolating between two detected sub-styles (as shown in Fig.~\ref{fig:sub-style transfer}). The mixing parameters $\beta_1$ and $\beta_2$ vary from 0 to 1 in 0.2 steps, i.e., $(\beta_1=0, \beta_2=1)$, $(\beta_1=0.2, \beta_2=0.8)$, $(\beta_1=0.4, \beta_2=0.6)$, etc., allowing the user to adjust the stylization process.}
	\label{fig:sub-style interpolation}
\end{figure*}

\begin{figure}[h]
	\centering

	\subfloat[Improved WCT (l4)~\cite{wynen2018unsupervised}]{\includegraphics[width=.22\textwidth]{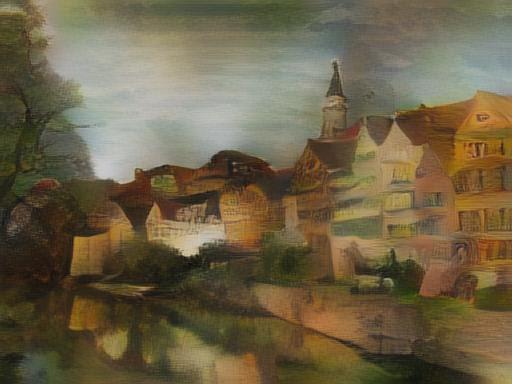}} \hspace{0.1px}	
	\subfloat[SST (l4) (proposed)]{\includegraphics[width=.22\textwidth]{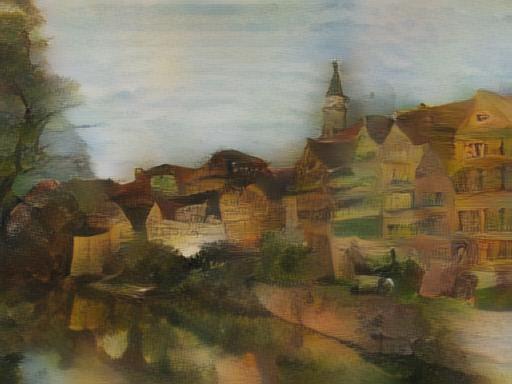}}

	\subfloat[Improved WCT (full)~\cite{wynen2018unsupervised}]{\includegraphics[width=.22\textwidth]{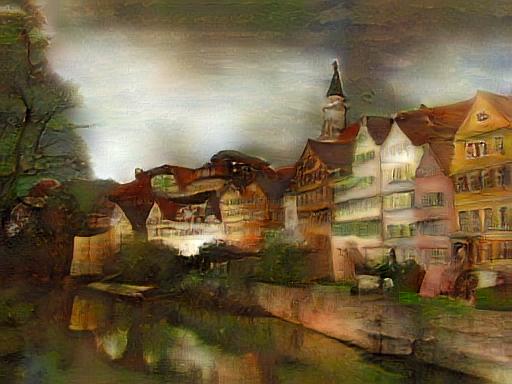}}  \hspace{0.1px}	
\subfloat[SST (full) (proposed)]{\includegraphics[width=.22\textwidth]{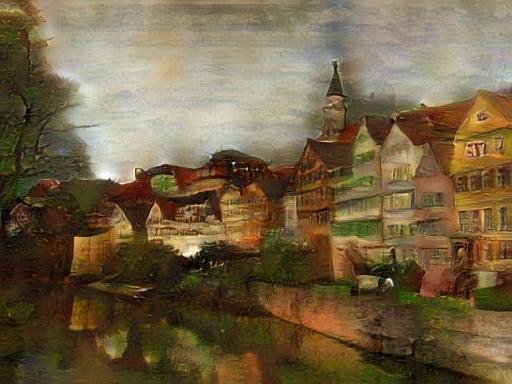}}

	\caption{Performing Semantic Sub-style Transfer (SST) on the fourth pooling layer of the VGG-19 network. The multi-level pipeline proposed in~\cite{li2017universal} was employed either using only the first two encoders, i.e., the fifth and fourth encoders (denoted by ``l4'), or using all the encoders (denoted by ``full'').  The content and style image are from Figure \ref{fig:masks}.  The style weight was set to $\alpha=0.6$, the content weight to $\delta=0.8$ and three sub-styles/sub-contents ($K=3$) were used  for all the conducted experiments. }
	\label{fig:sst-1}
\end{figure}

\begin{figure}[h]
	\center
	\subfloat[style 1]{\includegraphics[width = 0.25\linewidth]{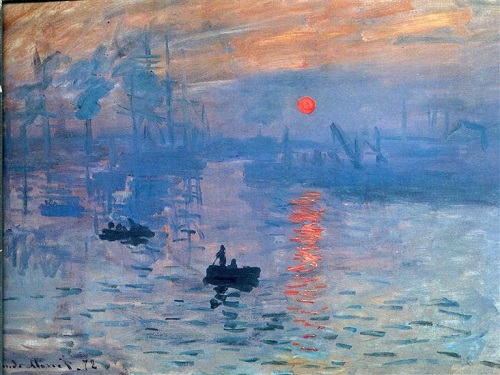}} 
	\subfloat[style 2]{\includegraphics[width = 0.25\linewidth]{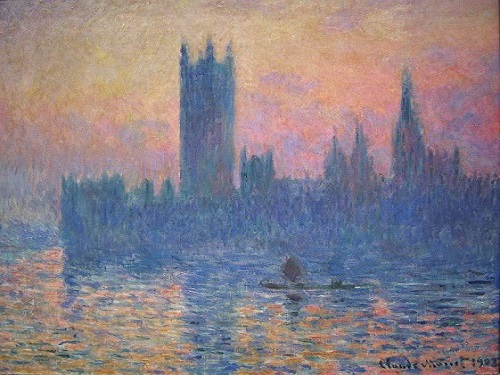}}
	\subfloat[style 3]{\includegraphics[width = 0.25\linewidth]{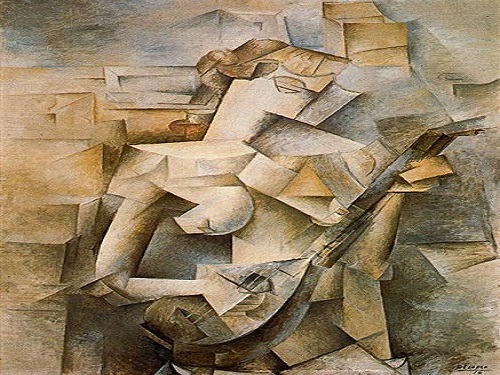}} 
	\subfloat[style 4]{\includegraphics[width = 0.25\linewidth]{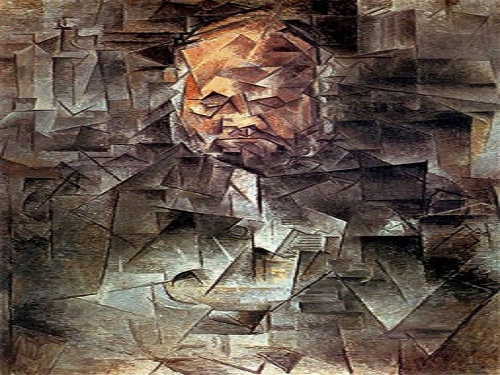}} \\
	\vspace{-0.8em}			
	\subfloat[content]{\includegraphics[width = 0.4\linewidth]{best1.jpg}}  \hspace{0.1px}	
	\subfloat[MST (all)]{\includegraphics[width = 0.4\linewidth]{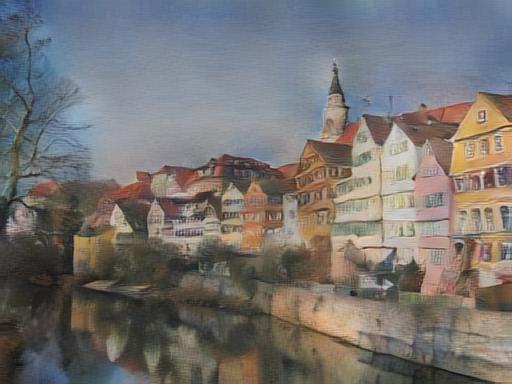}} 	\\
	\vspace{-0.8em}				
	\subfloat[MST (sub-style 1)]{\includegraphics[width = 0.4\linewidth]{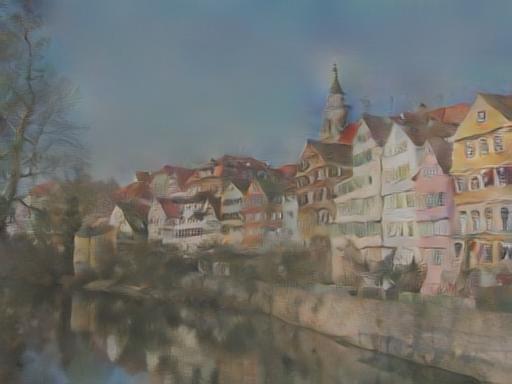}}  \hspace{0.1px}	
	\subfloat[MST (sub-style 2)]{\includegraphics[width = 0.4\linewidth]{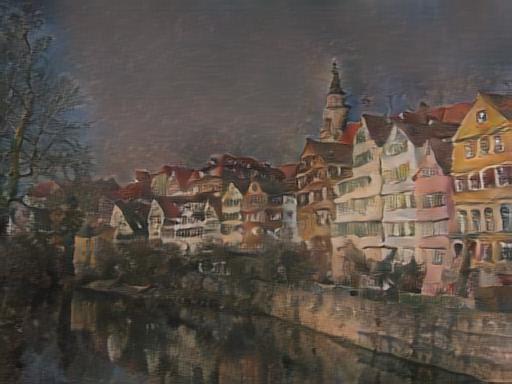}}

	\caption{Using the MST approach to detect and transfer two distinct sub-styles detected in a collection of style images. The images were stylized using the features extracted from the fourth pooling layer of the VGG-19 network.}
	\label{fig:mst}
\end{figure}

\begin{figure}[h]
	\center
\subfloat[Content]{\includegraphics[width = 0.45\linewidth]{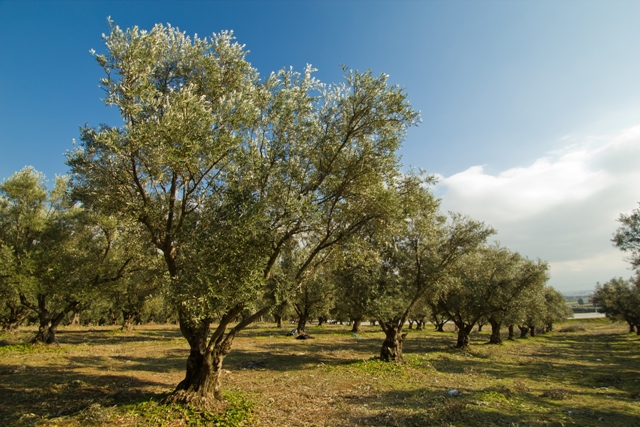}} \hspace{0.1px}
\subfloat[Style]{\includegraphics[width = 0.45\linewidth]{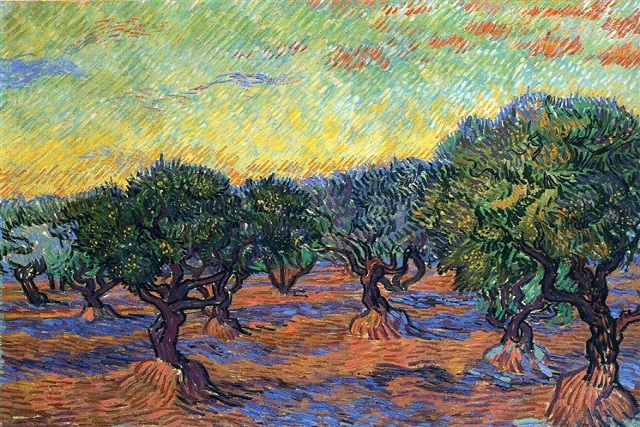}} \hspace{0.1px}\\
\vspace{-1em}
\subfloat[Improved WCT~\cite{wynen2018unsupervised}]{\includegraphics[width = 0.45\linewidth]{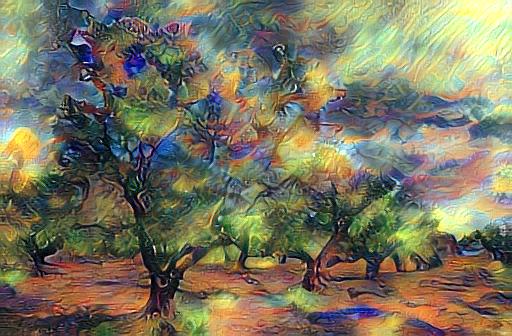}} \hspace{0.1px}
\subfloat[SST (proposed)]{\includegraphics[width = 0.45\linewidth]{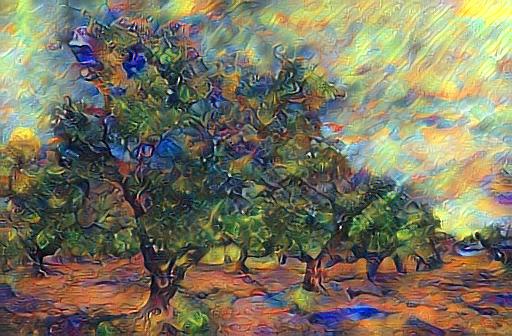}} \hspace{0.1px}\\

	\subfloat[Content]{\includegraphics[width = 0.45\linewidth]{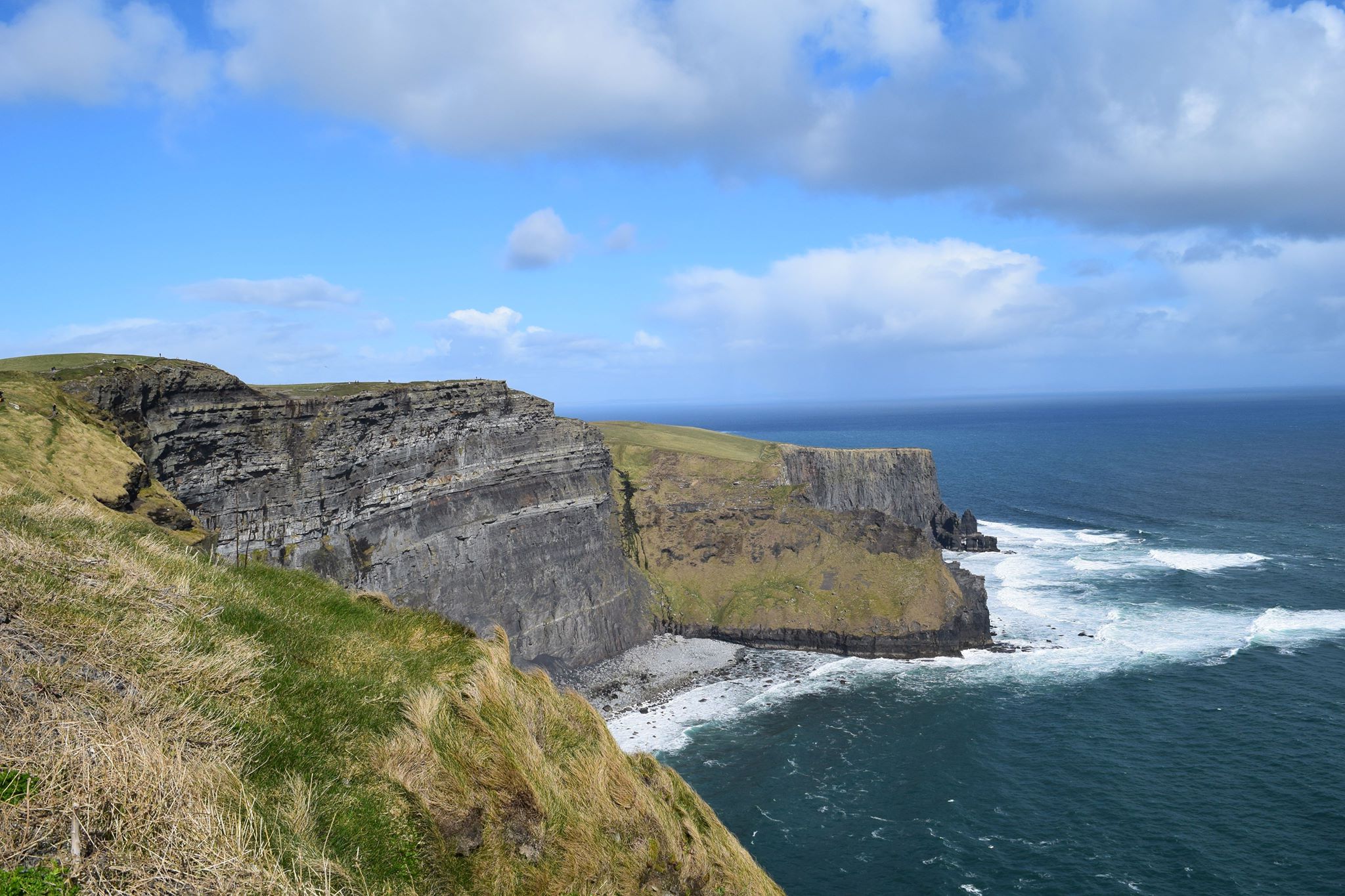}} \hspace{0.1px}
	\subfloat[Style]{\includegraphics[width = 0.45\linewidth]{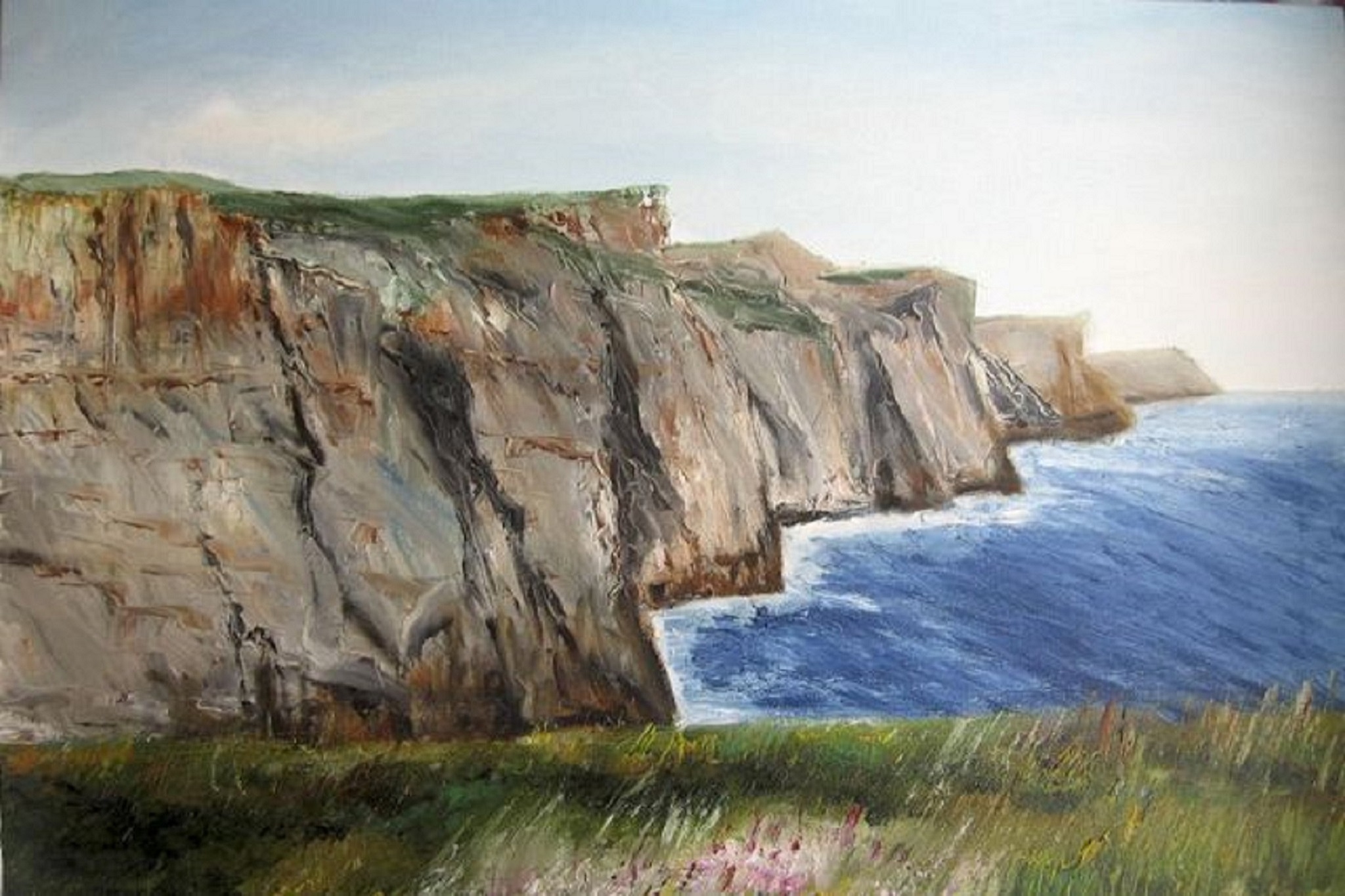}} \hspace{0.1px}\\
	\vspace{-1em}
	\subfloat[Improved WCT~\cite{wynen2018unsupervised}]{\includegraphics[width = 0.45\linewidth]{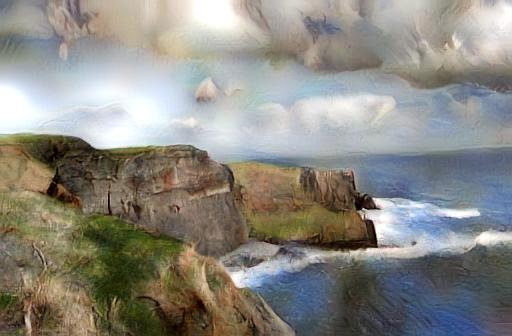}} \hspace{0.1px}
	\subfloat[SST (proposed)]{\includegraphics[width = 0.45\linewidth]{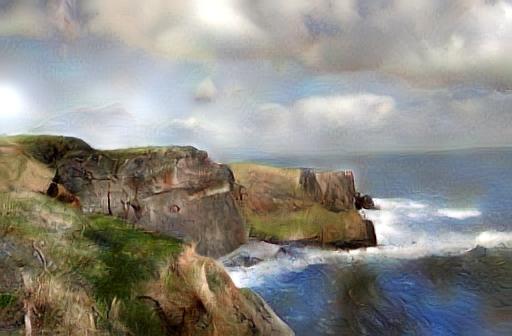}}\hspace{0.1px}\\

	\subfloat[Content]{\includegraphics[width = 0.45\linewidth]{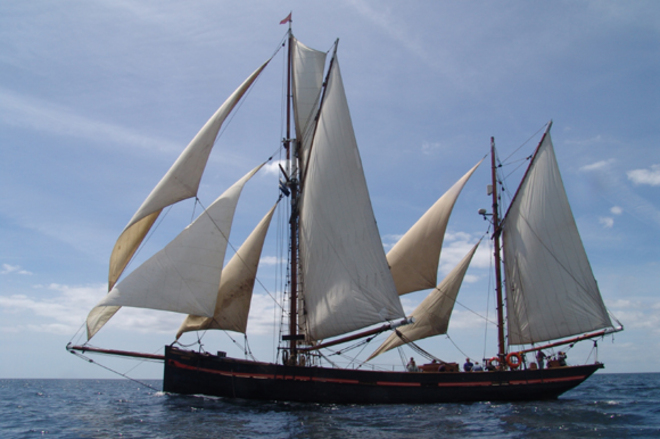}} \hspace{0.1px}
	\subfloat[Style]{\includegraphics[width = 0.45\linewidth]{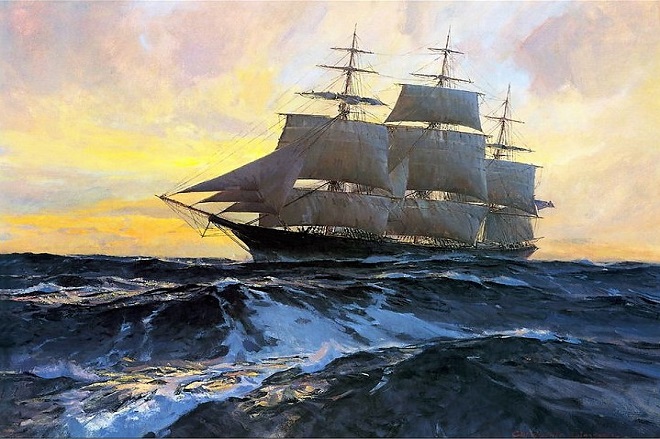}} \hspace{0.1px}\\
	\vspace{-1em}
	\subfloat[Improved WCT~\cite{wynen2018unsupervised}]{\includegraphics[width = 0.45\linewidth]{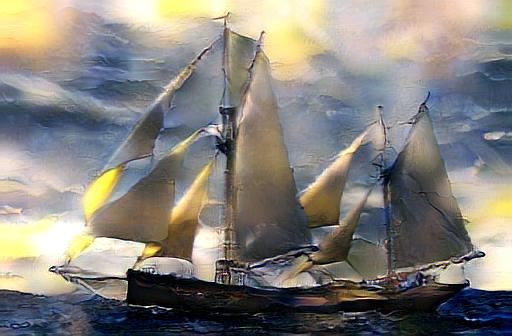}} \hspace{0.1px}
	\subfloat[SST (proposed)]{\includegraphics[width = 0.45\linewidth]{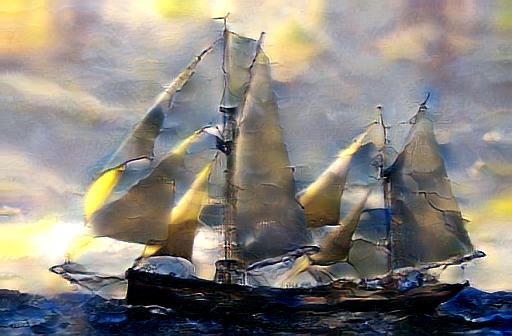}} \hspace{0.1px}\\

	\caption{Performing Semantic Sub-style Transfer (SST). The style weight was set to $\alpha$ = 0.6, while the content weight was set to $\delta = 0.8$ ($\delta = 1$ for the first image). The number of detected sub-styles was set to $K=6$ for the first image, $K=2$ for the second image and $K=4$ for the last one.}
	\label{fig:sst-2}
\end{figure}

\section{Experimental Evaluation}
\label{section:experimental}

For all the conducted experiments a VGG-19 network was employed for extracting the features from its 5 pooling layers (depending on the levels used for performing the stylization)~\cite{simonyan2014very}. Symmetric decoding networks, that employ nearest neighbor upsampling layers, were trained to reconstruct the input image using the features extracted from the various VGG-19 pooling layers. The networks were trained on the Microsoft COCO dataset~\cite{lin2014microsoft}. For all the conducted experiments, we used the WCT implementation provided by the authors of~\cite{wynen2018unsupervised}. Furthermore, the number of sub-content clusters $K_c$ was set to be equal to the number of sub-styles $K_s$, i.e., $K = K_c = K_s$. When the multi-level stylization procedure is applied~\cite{li2017universal}, all the five pooling layers of the VGG-19 network are used.

First, we examine whether it is indeed possible to detect different sub-styles within the same style image and use them to perform different stylizations of the content image. In Fig.~\ref{fig:sub-style transfer}, three different sub-styles were detected from each style image and each sub-style was separately transferred to the content image depicted in Fig.~\ref{fig:masks}. Note that the proposed SMT was indeed able to detect the variations of the style in the given style images, providing three significantly different ways to stylize a content image. On the other hand, the WCT approach is only capable of providing one fixed stylization. Then, the user can appropriately blend different sub-styles to achieve the aesthetic results that are desired, as demonstrated in Fig.~\ref{fig:sub-style interpolation}. Note that this approach provides a) the opportunity to better understand the different styles that exist within the same style image and b) the ability to adjust the stylization process to the needs of the user.

The previous experiments (Fig.~\ref{fig:sub-style transfer} and~\ref{fig:sub-style interpolation}) demonstrated that it is indeed possible to detect different sub-styles in a given style image and use them to perform different stylizations. However, we have not yet evaluated the ability of them to actually improve the process of NST, since they were merely used to perform different stylizations. The proposed SST method is capable of automatically using the detected sub-styles and matching them to the most appropriate regions of the content image in order to reduce the artifacts produced by the style transfer and improve the quality of the generated textures. This is illustrated in Fig.~\ref{fig:sst-1}, where the SST method was used to transfer the style using the image shown in Fig.~\ref{fig:masks}. The fourth pooling layer of the VGG-19 network was used for extracting the features used for the SST method, while $K=3$ sub-styles were used for the stylization. Then, multi-level stylization was employed either using only the first two encoders, i.e., the fifth and fourth encoders (denoted by ``l4'' in Fig.~\ref{fig:sst-1}), or using all the encoders (denoted by ``full''). Note that in both cases, the proposed method significantly reduces the artifacts, e.g., note the more uniform stylization of the sky, as well as improves the overall quality of the textures, e.g., note the more detailed texture of the houses. The improved behavior of the proposed approach is also confirmed in the examples provided in Fig.~\ref{fig:sst-2}.  For these experiments we employed the state-of-the-art blending approach proposed in~\cite{wynen2018unsupervised}. The multi-level stylization procedure was employed~\cite{li2017universal}, while the SST method was applied on the features extracted from the fourth pooling level of the VGG-19 network.

Finally, we evaluated the ability of the proposed method to detect consistent (sub-)styles over collections of style images (MST method). To this end, 4 styles images were provided, as shown in Fig.~\ref{fig:mst}, and the proposed method automatically detected two styles according to which the content image was stylized, following the same approach as in the previous experiments (Fig.~\ref{fig:sub-style transfer}). The two detected sub-styles indeed correspond to the two dominant styles that appear in the provided style images, with the first one being mainly influenced by style images 1 and 2, while the second one being influenced by style images 3 and 4. This is even more evident if we carefully examine the texture of the corresponding stylized images. This approach can be also used to study the evolution of the painting style of an artist through time, providing a useful visualization tool allowing for rendering the same image in different styles, as they would have been painted through the artist's life.

\section{Conclusions}
\label{section:conclusions}

Even though universal Neural Style Transfer (NST) methods are capable of performing style transfer of arbitrary styles in a style-agnostic manner, they usually fail to handle more complex styles that arise in more sophisticated style images. To overcome this limitation, we proposed a novel universal NST approach that separately models each sub-style that exists in a given style image (or a collection of style images), allowing for better modeling the sub-styles within the same style image and then use the most appropriate sub-style to stylize the content image. The flexibility of the proposed approach was demonstrated through extensive experiments.
The proposed approach is orthogonal to existing NST method, allowing it to be readily  combined with any of these approaches, further improving style transfer. Finally, whereas previous approaches aimed to merely improve style transfer, the proposed approach leads towards a better understanding of NST by decomposing the style of images, while giving the opportunity to the users to interfere with the final result, providing a practical NST tool.

\bibliographystyle{abbrv}
\bibliography{bib}

\end{document}